\documentclass[letterpaper, 10 pt, conference]{ieeeconf}  

\IEEEoverridecommandlockouts                              

\usepackage{booktabs}
\usepackage{cuted}
\usepackage{soul}
\usepackage{hyperref}
\usepackage{amsfonts} 
\usepackage{verbatimbox}
\usepackage{graphicx}
\usepackage[export]{adjustbox}
\usepackage{amsmath}
\usepackage{todonotes}
\usepackage{xcolor}
\usepackage{rotating}
\usepackage{afterpage}%
\usepackage{array}
\usepackage{caption,setspace}
\usepackage{subcaption}
\newcommand{\secref}[1]{Sec.~\ref{#1}}
\newcommand{\figref}[1]{Fig.~\ref{#1}}
\usepackage{letltxmacro}
\LetLtxMacro{\originaleqref}{\eqref}
\usepackage{pifont}
\newcommand{\cmark}{\ding{51}}%
\newcommand{\xmark}{\ding{55}}%

\newcommand{\tabref}[1]{Tab.~\ref{#1}}
\renewcommand{\eqref}{Eq.~\originaleqref}

\title{\LARGE \bf
Uncertainty-aware LiDAR Panoptic Segmentation
}

\author{Kshitij Sirohi$^{1}$, Sajad Marvi$^{1}$, Daniel B\"uscher$^{1}$ and Wolfram Burgard$^{2}$
\thanks{
$^{1}$Department of Computer Science at University of Freiburg, Germany.
$^{2}$Department of Engineering at Technical University Nürnberg, Germany.
This work was financed by the Baden-Württemberg Stiftung gGmbH.
}
}
\begin{document}
\maketitle
\thispagestyle{empty}
\pagestyle{empty}

\begin{abstract}
Modern autonomous systems often rely on LiDAR
scanners, in particular for autonomous driving scenarios. In
this context, reliable scene understanding is indispensable.
Current learning-based methods typically try to achieve maximum performance for this task, while neglecting a proper estimation of the associated uncertainties.
In this work, we introduce a novel approach for solving the task of uncertainty-aware panoptic segmentation using LiDAR point clouds.
Our proposed EvLPSNet network is the first to solve this task efficiently in a sampling-free manner.
It aims to predict per-point semantic and instance segmentations, together with per-point uncertainty estimates.
Moreover, it incorporates methods for improving the performance by employing the predicted uncertainties.
We provide several strong baselines combining state-of-the-art panoptic segmentation networks with sampling-free uncertainty estimation techniques.
Extensive evaluations show that we achieve the best performance on uncertainty-aware panoptic segmentation quality and calibration compared to these baselines.
We make our code available at: \url{https://github.com/kshitij3112/EvLPSNet}
\end{abstract}


\section{Introduction}

\begin{figure}
\captionsetup[subfigure]{aboveskip=1ex,belowskip=1ex}
\centering
\begin{subfigure}{0.9\linewidth}
 \includegraphics[width=0.9\linewidth]{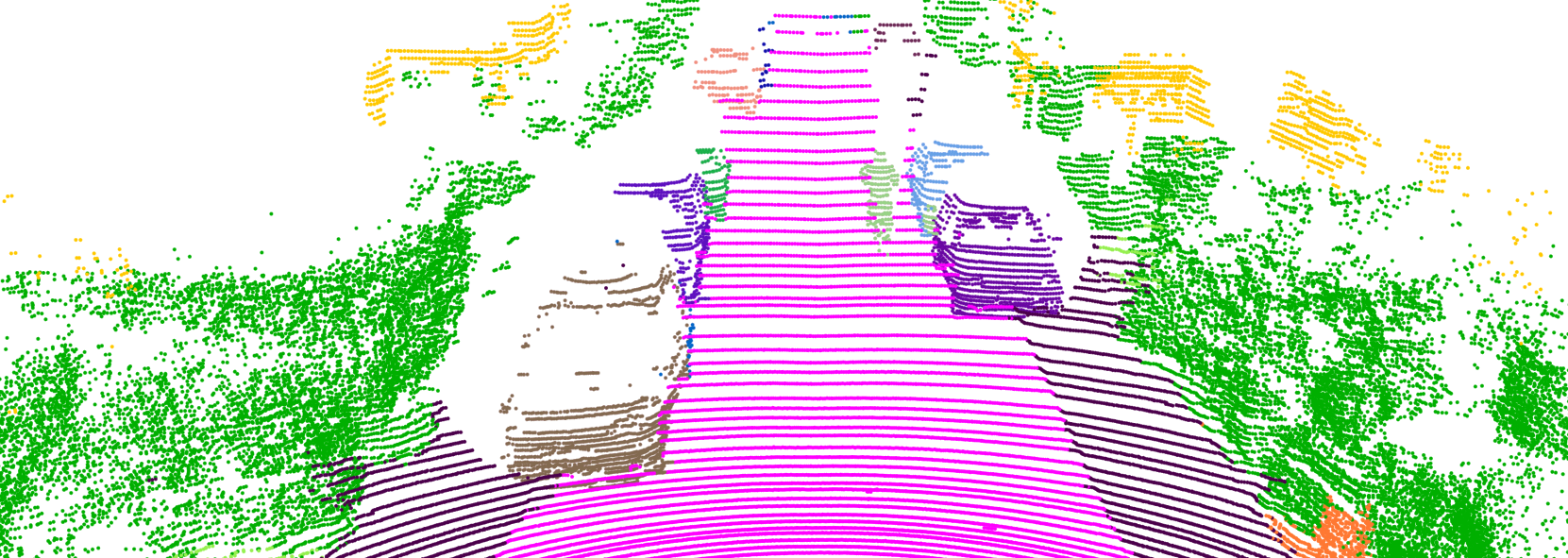}
 \subcaption{LiDAR panoptic segmentation}\label{fig:pan_seg}
\end{subfigure} 
\begin{subfigure}{0.9\linewidth}
 \includegraphics[width=0.9\linewidth]{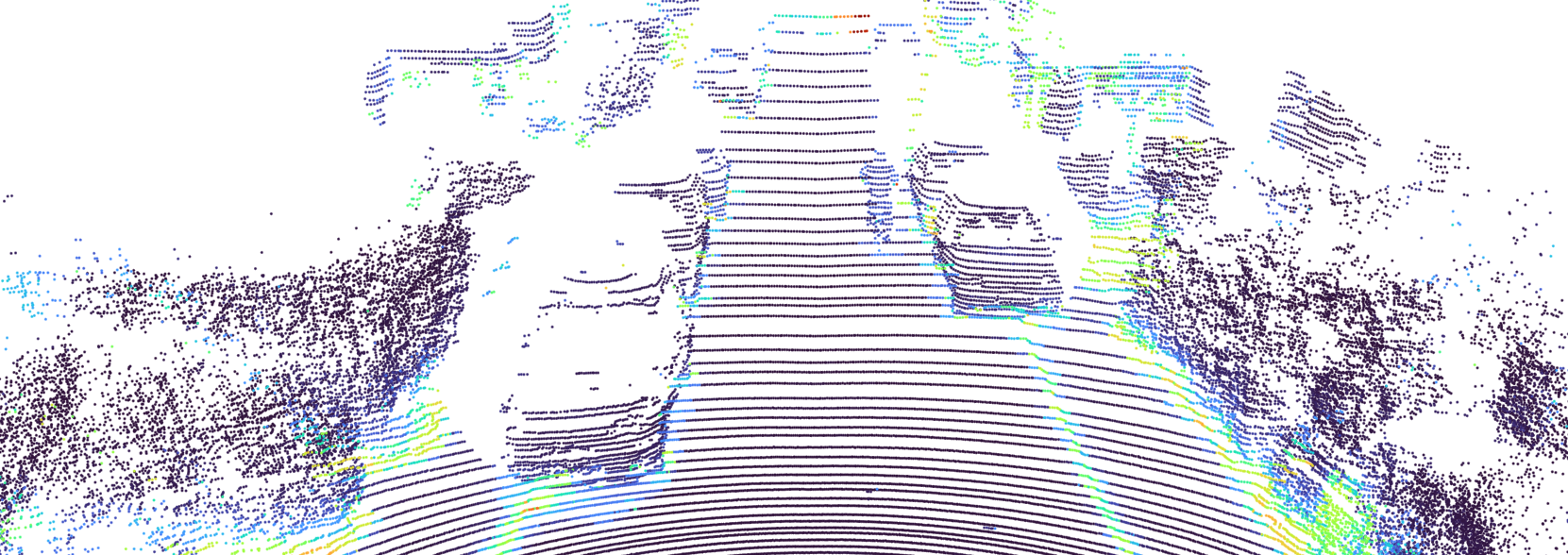}
 \subcaption{LiDAR panoptic uncertainties}\label{fig:pan_unc}
\end{subfigure}
\caption{Panoptic segmentation and associated uncertainties as predicted by our EvLPSNet for the SemanticKITTI validation dataset.}
\label{fig:examples}
\vspace{-0.5ex}
\end{figure}

A perception system capable of providing comprehensive and reliable scene understanding is crucial for the safe operation of an autonomous vehicle.
The recently introduced panoptic segmentation \cite{kirillov2019panoptic} unifies the semantic segmentation of \textit{stuff} and instance segmentation of \textit{thing} classes into a single task.
This facilitates the evaluation of the overall accuracy, which is crucial for a holistic scene understanding.
In practice, however, the performance can only be evaluated on a limited dataset, while the real-world consists of scenarios and objects possibly not present in the dataset.
Therefore, in addition to an evidence signal, a reliable uncertainty estimate is crucial for safety-critical applications, such as autonomous driving.
Hence, the task of uncertainty-aware panoptic segmentation \cite{sirohi2022uncertainty} for a unified evaluation of panoptic segmentation and uncertainty estimation
offers a better potential for deployment.
Our method aims to solve this task for LiDAR point-clouds, as illustrated in \figref{fig:examples}.

The regular grid structure of images allows a number of works on the panoptic segmentation
to take advantage of recent advances in deep learning, in particular using convolutional neural networks (CNNs) \cite{mohan2021efficientps,cheng2020panoptic}.
On the other hand, the irregular, sparse and unordered structure of LiDAR point clouds posed unique challenges.
However, LiDARs provide an illumination-independent accurate geometric description of the environment,
yielding a great advantage over images.
This motivated recent works for panoptic segmentation of LiDAR point clouds,
represented in various ways,
such as range images \cite{sirohi2021efficientlps,milioto2020lidar,triess2020scan}, 3D voxels \cite{yan2018second},
birds-eye-views (BEVs) \cite{zhang2020polarnet}, or direct points \cite{thomas2019kpconv}.
These methods are generally classified into proposal-based \cite{sirohi2021efficientlps} and proposal-free \cite{zhou2021panoptic}.

Conventional CNN-based methods, utilizing the softmax operation, typically show overconfidence in their predictions \cite{sensoy2018evidential}.
On the other hand, popular sampling-based methods for uncertainty estimate methods,
such as Monte Carlo dropout \cite{gal2016dropout}, and Bayesian neural networks (BNNs) \cite{kendall2015bayesian},
are time and memory intense hence not suitable for real-time applications.
Therefore, there is a recent interest in sampling-free methods for uncertainty estimation,
such as evidential deep learning \cite{sensoy2018evidential}, predicting uncertainties in a single pass.
However, most of these works for classification or segmentation deal with the image domain.
Hence, to the best of our knowledge, there is still no existing approach to provide sampling-free point-wise uncertainty estimates for the panoptic segmentation of LiDAR point clouds. 

In this work, we present the novel Evidential LiDAR Panoptic Segmentation Network (EvLPSNet),
the first network to tackle this task, by utilizing evidential deep learning.
We use the 2D polar BEV grid representation \cite{zhou2021panoptic} for our network, facilitating fast inference times and better separability of instances.
However, the projection into the grid structure leads to discretization errors, as all points in a grid cell are assigned the same prediction.
We approach this issue using the 3D point information, as well as our uncertainty estimates,
proposing a novel learnable uncertainty-based Query and Refinement (uQR) module.
This module employs a simple point-based convolution layer to achieve point-wise predictions for points selected based on their uncertainty.
We also propose to utilize the predicted probabilities to create an efficient version of the k nearest neighbors algorithm (pKNN).
Furthermore, we provide several baselines and evaluate their results on the task of uncertainty-aware LiDAR panoptic segmentation. 
In summary, our contributions are as follows:
\begin{itemize}
    \item The novel proposal-free EvLPSNet architecture for uncertainty-aware LiDAR panoptic segmentation.
    \item The uQR module for refining the prediction for the most uncertain points.
    \item The efficient pKNN algorithm utilizing the predicted class probabilities.
    \item Several baselines for comparison with EvLPSNet.
\end{itemize}

\section{Related Work}

\subsection{Segmentation of LiDAR Point Clouds}

The release of the SemanticKITTI dataset \cite{behley2019semantickitti} led to the emergence of many works, initially for the semantic segmentation of LiDAR point clouds.
These can generally be classified based on the point cloud representations they employ,
such as projected range images \cite{milioto2019rangenet++, cortinhal2020salsanext, kochanov2020kprnet},
3D voxels \cite{tang2020searching}, point-based \cite{thomas2019kpconv}, and BEV polar coordinates \cite{zhang2020polarnet}.
Most panoptic approaches utilize these representations as well.

Panoptic segmentation approaches can be classified as proposal-based and proposal-free.
While both employ separate semantic and instance segmentation branches, the distinction lies in the latter.
Proposal-based methods typically employ bounding box regression for discovering instances,
such as Mask-RCNN \cite{he2017mask} in the case of EfficientLPS \cite{sirohi2021efficientlps}.
On the other hand, proposal-free approaches perform clustering on the semantic prediction to obtain instance ids for objects belonging to separate instances.
Panoptic-PolarNet \cite{zhou2021panoptic} utilizes a Panoptic Deeplab-based \cite{cheng2020panoptic} instance head to regress offsets and centers for different instances.
DS-Net \cite{zhao2022divide} proposes a dynamic shifting module to move instance points towards their respective center.
Panoptic-PHNet \cite{li2022panoptic} utilizes two different encoders, BEV and voxel-based, to encode point cloud features,
followed by a KNN-transformer module to model interaction among voxels belonging to thing classes.

\subsection{Uncertainty Estimation}

Many works for estimating uncertainty in segmentation tasks employ sampling-based methods, such as Bayesian Neural Networks \cite{kendall2015bayesian} or Monte Carlo dropout \cite{gal2016dropout,huang2018efficient}.
However, such methods are time and memory-intensive, requiring multiple passes or sampling operations.
For LiDAR point clouds, SalsaNext \cite{cortinhal2020salsanext} is an uncertainty-aware semantic segmentation utilizing BNNs.
Even though the network output is quick to evaluate, due to the sampling of the BNN approach the uncertainty is slow to obtain.
Further, no metric is presented to quantify the calibration of the predicted uncertainty for this approach.
We believe these are severe limitations for safety-critical real-time applications like autonomous driving.
The need for single-pass sampling-free uncertainty estimation motivates many works in the field.
Classical neural networks utilize softmax operations of the final logits to predict per class score or probability,
which is not a reliable estimate of the network's confidence in the prediction, as shown by \cite{sensoy2018evidential}.
Guo et al. \cite{guo2017calibration} propose the Temperature Scaling (TS) method
to learn a logit scaling factor on the softmax operation to provide calibrated probability predictions.
Other methods, such as \cite{li2022uncertainty}, learn to separate different classes in a latent space and,
based on the distance of the predicted to the nearest class feature, calculate the uncertainty.

Sensoy et al. \cite{sensoy2018evidential} proposed evidential deep learning
to provide reliable and fast uncertainty estimation with minimal changes to a network.
Petek et al. \cite{petek2022robust} utilize this method to simultaneously predict semantic segmentation and bounding box regression uncertainty.
Sirohi et al. \cite{sirohi2022uncertainty} introduce the uncertainty-aware panoptic segmentation task
and provide a sampling-free network for a unified panoptic segmentation and uncertainty for images.
In our present work, we build upon this to extend the approach to LiDAR point-clouds
and we provide a comprehensive quantitative analysis.

\begin{figure*}
\centering
\includegraphics[width=0.9\textwidth]{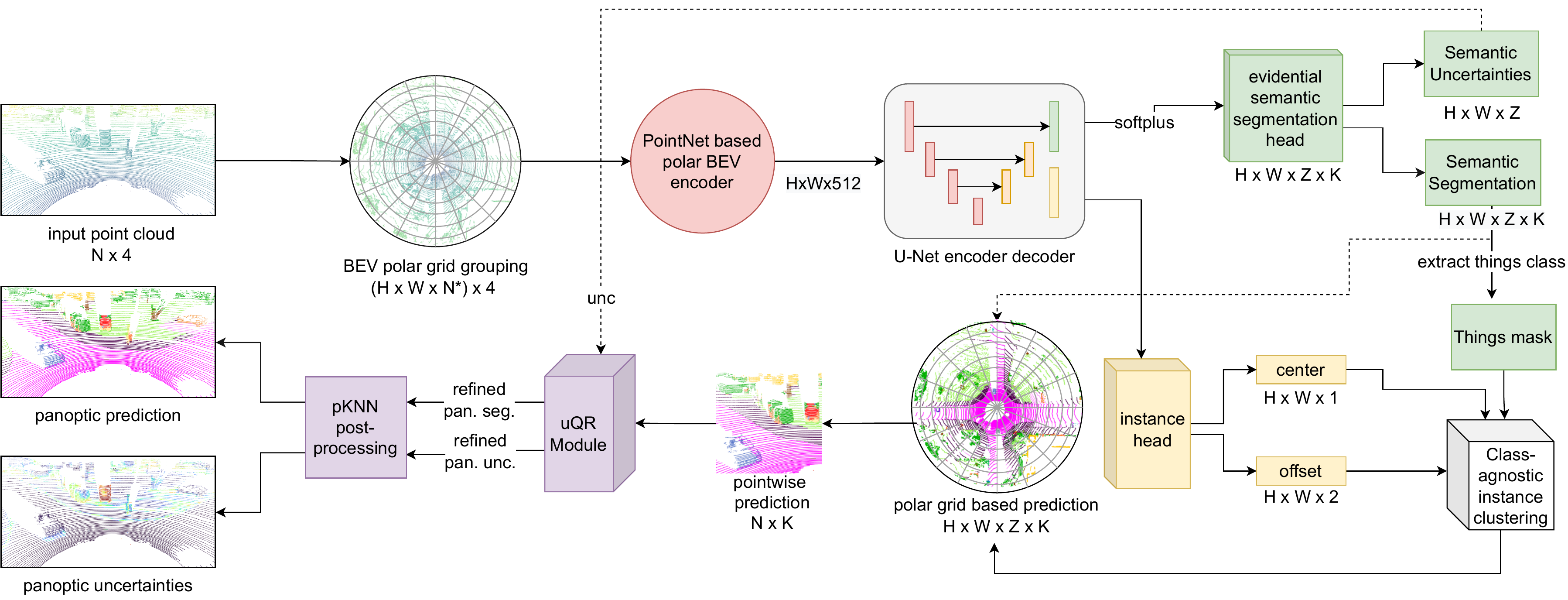}
\caption{Overview of our EvLPSnetwork architecture.}
\label{fig:network}
\vspace{-0.5ex}
\end{figure*}

\section{Technical Approach}

An overview of our network architecture is shown in \figref{fig:network}.
It is based on the proposal-free Panoptic-PolarNet network \cite{zhou2021panoptic}.
Our evidential semantic segmentation head and Panoptic-Deeplab based \cite{cheng2020panoptic} instance segmentation head
utilize the learned features to predict per-point semantic segmentation, semantic uncertainty, instance center and offsets.
The predictions from both heads are fused to provide panoptic segmentation results.
Leveraging the segmentation uncertainties, our proposed query and refine module helps to improve the prediction for points within uncertain voxels.
Moreover, post-processing using our efficient probability-based KNN improves the results further.

\subsection{Network Architecture}

We project the LiDAR points into a polar BEV grid utilizing the encoder design proposed by PolarNet \cite{zhang2020polarnet}.
First, the points (represented in 3D polar coordinates) are grouped according to their location in a 2D polar BEV grid.
The grid has the dimensions of $H \times W = 480 \times 360$, where $H$ corresponds to the range and $W$ to the heading angle.
Then, for each grid cell, the corresponding points are encoded using a simplified PointNet \cite{qi2017pointnet}.
This is followed by a max pooling operation to calculate the feature vector for every 2D grid cell and to create a fixed-size grid representation of $W \times H \times F$,
where $F=512$ is our number of feature channels.

The subsequent encoder-decoder network utilizes the U-net \cite{ronneberger2015u} architecture.
Its first three decoder layers are shared by the semantic and instance segmentation branches, while the remaining layers are separate.
The instance segmentation regresses the instance center heatmap and the instance offsets on the BEV grid.  

\subsection{Evidential Semantic Segmentation}

We utilize evidential deep learning \cite{sensoy2018evidential} to provide voxel-level semantic segmentation with calibrated uncertainty estimation.
Here, the voxels are in polar coordinates with dimension $H \times W \times Z$, where $Z = 32$ corresponds the vertical segmentation of the BEV grid.
Our network is based on Panoptic-PolarNet \cite{zhou2021panoptic}.

To incorporate  the uncertainty estimate, we first add a softplus activation function on the final logits of the network, which works as the evidence signal,
essentially signifying the evidence or weight collected for a particular class.
Then instead of treating the per class prediction as a single-value estimate, we employ the Dirichlet distribution for our per-point multinomial classification  \cite{sensoy2018evidential}.
For each point $i$, the Dirichlet distribution is parametrized by $\alpha = [\alpha^{1}, ..., \alpha^{K}]$,
where $K$ is the number of classes and $\alpha^k_i = \text{softplus}(l^k_i) + 1$ for network logit output $l^k_i$ for class $k$.
The corresponding probability $p_i$ and uncertainty $u_i$ are calculated as:

\begin{align}
p^k_i &= \alpha^k_i / S_i \label{prob_equation} \\
u_i &= K / S_i\label{unc_equation},
\end{align}
where $S_i = \sum_{k=1}^K  \alpha^k_i$.

We train the semantic segmentation head using the type-II maximum likelihood version of the loss and the KL term, provided by \cite{sensoy2018evidential},
to regularize the evidence magnitude in order to predict high uncertainty for wrongly predicted classes.
The loss is given by:
\begin{equation}
\mathcal{L}_\text{sem} = \mathcal{L}_{\log}^\text{s} + \lambda_t \mathcal{L}_\text{KL}^\text{s},
\end{equation}
We use $\lambda_t = 0.065 \times \text{min}\{1, t/(20I)\}$, where $t$ is the current training iteration and $I$ is the number of iterations per epoch. Thus $\lambda_t$ increases until epoch 20 and then remains 0.055. The log loss is given as:
 \begin{equation}\label{eq:log_loss_sem}
\mathcal{L}_{\log}^\text{s} = \sum_{i=1}^N\sum_{k=1}^K o^k_i \log(S_i / \alpha^c_i).
\end{equation}
where $N = W \cdot H \cdot Z$ is the total number of voxels, and $o$ is the one-hot encoded vector, which is 1 for the ground truth class and 0 otherwise.

In our experiments, we found that the performance stagnates after some point due to a high number of empty voxels.
On the other hand, if we only use the occupied voxels ($N=$ number of occupied voxels in \eqref{eq:log_loss_sem}), the uncertainty estimation results are not calibrated.
Hence, we first train the network with all voxels, and then, after performance convergence, we train for some epochs with only the occupied voxels.
This improves both the segmentation performance and uncertainty calibration.

\subsection{Instance Segmentation}

Similar to \cite{zhou2021panoptic}, we base our instance segmentation head on Panoptic-Deeplab \cite{cheng2020panoptic}. 
The instance segmentation head consists of separate center prediction and offset prediction heads. 
The former predicts the likelihood of each grid cell being the center of an instance, 
while the latter predicts a 2D offset for each grid cell to its center in polar coordinates. 
We encode the ground truth heatmap as a 2D Gaussian around the center of each instance.

Note that the instance predictions operate in the 2D BEV domain, which allows the application of well-researched and fast 2D convolution operations.
Another benefit of the BEV is the easy separation of objects that are close/overlapping in the heading or elevation angles, which is not the case in the range projection.

\subsection{Panoptic Segmentation}

To obtain the panoptic segmentation we are following \cite{zhou2021panoptic},
by first extracting the top $k$ centers of the instance segmentation after applying non-maximum suppression.
Next, we utilize the semantic segmentation to create a foreground mask,
where at least one of the $Z$ predictions in a BEV cell belongs to a thing class.
We group the objects and assign an instance id $I$ in the foreground mask based on their distance to the nearest of the $k$ centers.
Then, we assign the instance id $I$ to the thing class predictions in the semantic segmentation output.
Finally, we assign the instance class label based on a majority voting \cite{zhou2021panoptic} utilizing the evidential probabilities within the same instance group.
All the points belonging to the stuff class get their class label from the semantic segmentation.

\begin{figure}
\centering
\includegraphics[width=0.9\linewidth]{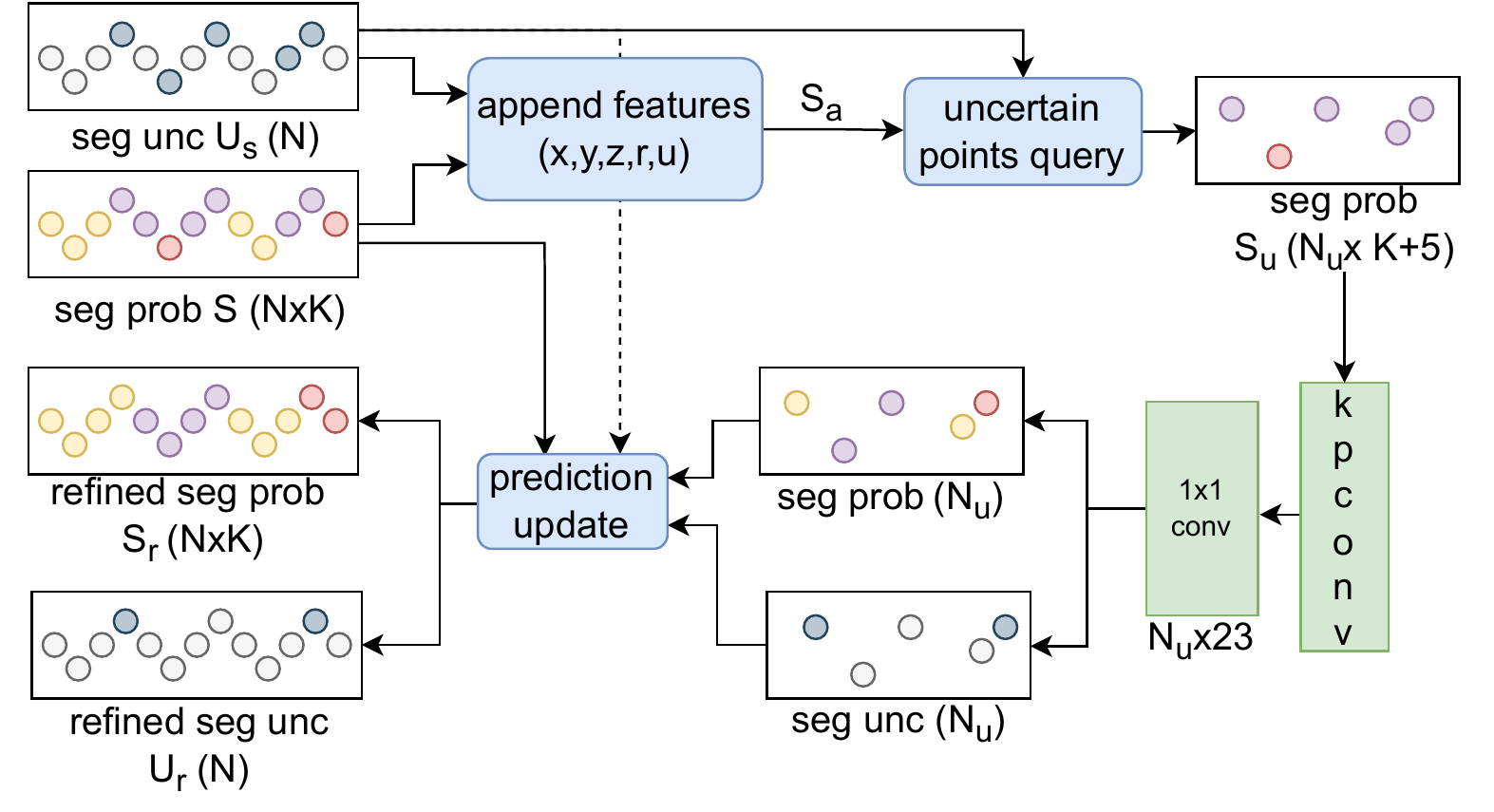}
\caption{Our uncertainty-based query and refinement module (uQR).} 
\label{fig:uQR}
\vspace{-0.5ex}
\end{figure}

\subsection{Uncertainty-based Query and Refinement module}

Our uncertainty-based query and refinement module (uQR) leverages the predicted uncertainties to counter the discretization errors due to the BEV grid structure.
We select the top $N = 20$k most uncertain points and pass them to our uQR module to actively improve the segmentation quality in an efficient way.
An overview of the module is shown in \figref{fig:uQR}.
While the semantic segmentation head makes voxel-wise predictions (which are simply transferred to the corresponding points), the uQR module does refined point-wise predictions. 

For $N$ points and $K$ classes, let $S \in \mathbb{R}^{N\times K}$ be the semantic segmentation probabilities calculated using \eqref{prob_equation}
and $U_s \in \mathbb{R}^{N}$ be the associated uncertainties from \eqref{unc_equation}.
First, we append each point by its $x,y,z$ coordinates, remission value, and uncertainty value to obtain $S_a\in \mathbb{R}^{N\times K+5}$ with a feature size of $K+5$.
Then we create a subset $S_u \in \mathbb{R}^{N_u\times K+5}$ of the $N_u$ most uncertain points and pass these to the KPconv-based network \cite{thomas2019kpconv} for refining.
KPconv utilizes point-based convolution in 3D space by capturing contextual information from its neighbors.
We only utilize one KPConv layer, followed by ReLU and a final classifier layer to
obtain the refined predictions and uncertainties as $S_{r} \in \mathbb{R}^{N\times K}$ and $U_{r} \in \mathbb{R}^{N}$.

\subsection{Efficient probability-based KNN}

In this section we devise our efficient probability-based k nearest neighbors (pKNN) approach.
Post-processing methods based on KNNs can be employed to cluster the instances or improve the segmentation quality.
However, their the speed of execution is a typical limitation.
We try to decrease the execution time by limiting the number of points requiring post-processing
by taking advantage of the predicted probabilities.
We first select points that have a probability (\eqref{prob_equation}) below a certain threshold. 
Then we find its $k$ nearest neighbors in the whole point cloud followed by a majority voting to decide the final label.
We choose $k=5$ and a probability threshold of $0.4$.
If a label is transferred from a neighboring point, the we corresponding uncertainty is transferred as well.

\section{Experimental Evaluation}

We evaluate the performance of our network on the challenging SemanticKITTI \cite{behley2021benchmark} dataset.
The dataset consists of 43,551 LiDAR scans distributed over 21 driving sequences.
Sequences 00 to 10 are used for training, except the sequence 08 which is reserved for validation, and sequences 11 to 21 are for testing.
The dataset provides point-wise annotation for 20 classes, out of which 8 are \textit{thing} classes and contain unique instance ids.
As the test set annotations are not provided, we evaluate the uncertainty-aware panoptic segmentation on the validation set.

\subsection{Baselines}

We aim to tackle the task of uncertainty-aware LiDAR panoptic segmentation for autonomous driving scenarios.
Hence, the uncertainty estimation method should be sampling-free and not create extra computation overhead.
The temperature scaling (TS) \cite{guo2017calibration} and evidential learning (Ev) \cite{sensoy2018evidential} qualifies both criteria.
For panoptic segmentation, we choose the proposal-based EfficientLPS \cite{sirohi2021efficientlps} and the proposal-free Panoptic-PolarNet \cite{zhou2021panoptic}.
We train these networks to the best setting provided by the official codes.
However, for a fair comparison we do not use pseudo labels for EfficientLPS.
First, we evaluate both original networks without any uncertainty estimation method involved.
Then, for temperature scaling, we add a scaling parameter to the semantic segmentation logits,
freeze the networks and train the scaling parameter until it converges on the validation set,
as suggested by the original authors.
Finally, we train our network utilizing evidential learning, our proposed uncertainty-based query and refinement module (uQR),
and our efficient probability-based KNN (pKNN) post-processing.

\subsection{Training Procedure}

We discretize the space into the grid size of $480 \times 360 \times 32$ polar voxels within the range $r \in  [3, 50]$\,m and height $z \in [-3, 1.5]$\,m w.r.t. the LiDAR scanner.
For non-maximum suppression, we use kernel size of 5 with 0.1 threshold and select top $k=100$ centers similar to \cite{zhou2021panoptic}.
We train the network for 50 epochs on single NVIDIA TITAN RTX GPU with a batch size of 3.
We use Adam optimizer with step learning rate with an initial value of 0.01 and a drop by a factor of 10 at epoch 40 and 45.

We apply instance augmentation and random data flipping along the $x$ and $y$ axes as suggested by \cite{zhou2021panoptic}.
However, we do not utilize their Self Adversarial Pruning (SAP) since it requires two forward passes per iteration, slowing down the training significantly.
In contrast, we utilize Lov{\'a}sz Evidential loss \cite{sirohi2022uncertainty} ($\mathcal{L}_\text{lev}$) for the last five epochs, which significantly improves the performance.
Further, we employ the evidential loss for semantic segmentation $\mathcal{L}_\text{sem}$ from \eqref{eq:log_loss_sem},
MSE loss ($\mathcal{L}_\text{h}$) for the center heatmap and
L1 loss ($\mathcal{L}_\text{o}$) for the offset.
The overall loss is
\begin{align}
    \mathcal{L} = \mathcal{L}_\text{sem} + \lambda_\text{h} \mathcal{L}_\text{h} + \lambda_\text{o} \mathcal{L}_\text{o} + \lambda_\text{lev} \mathcal{L}_\text{lev},
\end{align}
where $\lambda_\text{h} = 100$, $\lambda_\text{o} = 10$ and $\lambda_\text{lev} = 1$ for the last five episodes and $\lambda_\text{lev} = 0$ otherwise.
After the main network, we train the uQR module with
$\mathcal{L}_\text{sem} + \mathcal{L}_\text{lev}$
for 15 epochs with the Adam optimizer and a learning rate of 0.0001.

\subsection{Metrics}

We evaluate the performance of our network based on the uncertainty-aware Panoptic Quality (uPQ) and panoptic Expected Calibration Error (pECE) metrics as proposed in \cite{sirohi2022uncertainty}.
The metrics were utilized for evaluating performance for images where each instance consists of many densely organized pixels.
However, LiDAR point clouds are sparse, and separate instances generally contain fewer points, which can lead to biases in the metrics.
Hence, we adjusted the metrics to fit the sparse LiDAR points setting.

First, we search for the unique matching pairs between the ground truth and the prediction having IoU $>$ 0.5.
For each matching pair we calculate the average accuracy (acc) and confidence (conf).
Similar to \cite{sirohi2022uncertainty}, we calculate the confidence as $\text{conf}_i = 1 - u_i$, where $u_i$ is the predicted uncertainty for point $i$.
Further, we define acc $=$ 1 if the predicted class and instance id matches the ground truth and acc $=$ 0 otherwise.
Then, we average the accuracy in $J=10$ bins of confidence (from 0 to 1) for each class separately over the full dataset.
This is opposed to \cite{sirohi2022uncertainty},
where this calculation was done separately for each instance.
Finally, we calculate uECE$_k$ for each class $k$, pECE and uPQ as:
 \begin{align}
\text{uECE}_k &= \sum_{j=1}^{J} \frac{|B_j|}{N} \left| \text{acc}(B_j) - \text{conf}(B_j) \right| \\
\text{pECE} &= \frac{1}{K} \sum_k \text{uECE}_k \\
\text{uPQ} &= (1-\text{pECE}) \text{PQ}.
\end{align}
where $|B_{j}|$ is the number of points in bin $j$, and acc($B_j$) and conf($B_j$) are the average accuracy and confidence for bin $B_j$.
In addition, we provide separate results for the \textit{thing} and \textit{stuff} classes.
Further, we provide the mean Intersection over Union (mIoU) and the semantic uECE to evaluate the semantic segmentation and uncertainty estimation performance.

\begin{table*}
\begin{center}
\rule{0pt}{0.5ex}    
\caption{Performance values in \% on the SemanticKITTI validation set. Lower values are better for $\downarrow$, and larger values otherwise.}
\label{tab:quant_val}
\footnotesize
\begin{tabular}
{l|ccc|ccc|ccc|cc}
\toprule

Method & uPQ & PQ  & pECE $\downarrow$ & uPQ\textsuperscript{Th}& PQ\textsuperscript{Th} & pECE\textsuperscript{Th} $\downarrow$ & uPQ\textsuperscript{St}  & PQ\textsuperscript{St} & pECE\textsuperscript{St} $\downarrow$ &uECE $\downarrow$ & mIOU\\
\midrule
EfficientLPS &  $48.7$  & $57.1$ & $14.6$ & $52.6$  & $59.9$ & $\textbf{12.1}$ &  $45.9$ & $\textbf{54.9}$ & $16.4$ & $16.5$ & $62.3$ \\

EfficientLPS + TS  &   $49.6$ & $56.9$ & $12.9$ & $48.5$ & $59.8$ & $18.8$ &$50.2$& $54.9$ & $8.7$  & $7.7$ & $62.1$ \\
Panoptic-PolarNet  & $48.8$ & $\textbf{58.5}$ & $16.6$ & $\textbf{53.1}$ & $\textbf{65.7}$ & $19.1$ & $45.4$ & $53.3$ & $14.7$ & $13.2$ & $63.2$ \\
Panoptic-PolarNet + TS & $48.1$ & $\textbf{58.5}$ & $17.7$ & $51.4$ & $65.7$ & $21.8$ & $45.4$ & $53.3$ & $14.7$ & $10.5$ & $63.3$   \\
\midrule
EvLPSNet & $\textbf{51.4}$ & $58.0$ & $\textbf{11.5}$ & $52.7$ & $62.7$& $15.9$ & $\textbf{50.1}$ & $54.6$ & $\textbf{8.2}$ & $\textbf{7.1}$ & $\textbf{64.0}$ \\
\bottomrule
\end{tabular}
\end{center}
\end{table*}

\begin{figure*}
\centering
\footnotesize
\setlength{\tabcolsep}{0.05cm}
{
\renewcommand{\arraystretch}{0.2}
\newcolumntype{M}[1]{>{\centering\arraybackslash}m{#1}}
\begin{tabular}{cM{0.3\linewidth}M{0.3\linewidth}M{0.3\linewidth}}
& Panoptic Segmentation & Uncertainty Map & Error Map \\
\\
\\
\\
\rotatebox[origin=c]{90}{(a)} & {\includegraphics[width=\linewidth, frame]{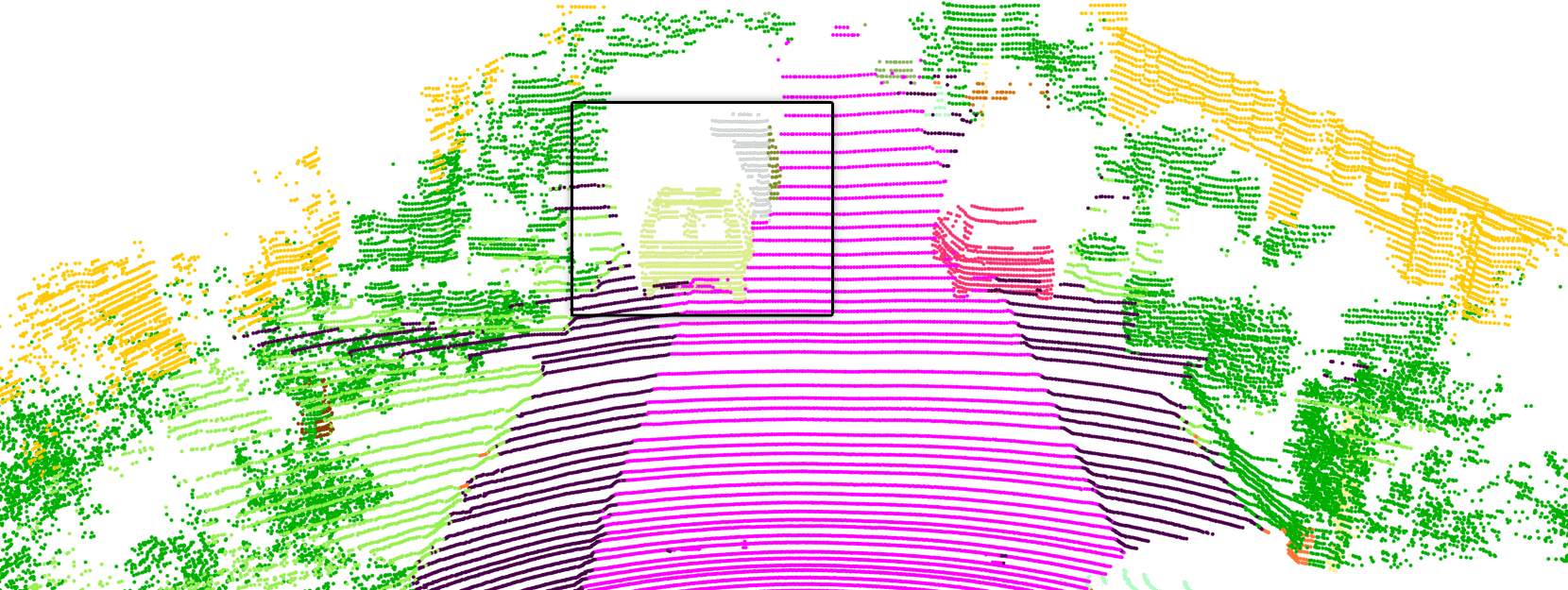}} & {\includegraphics[width=\linewidth, frame]{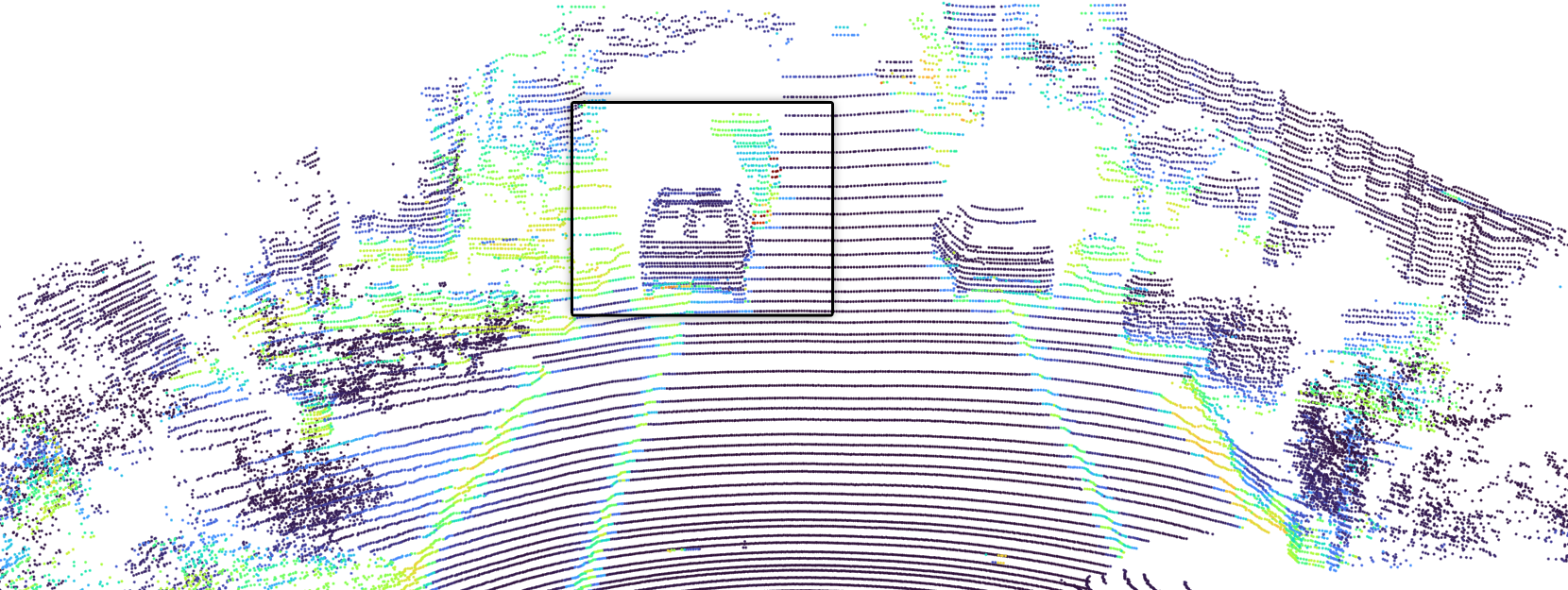}} & {\includegraphics[width=\linewidth, frame]{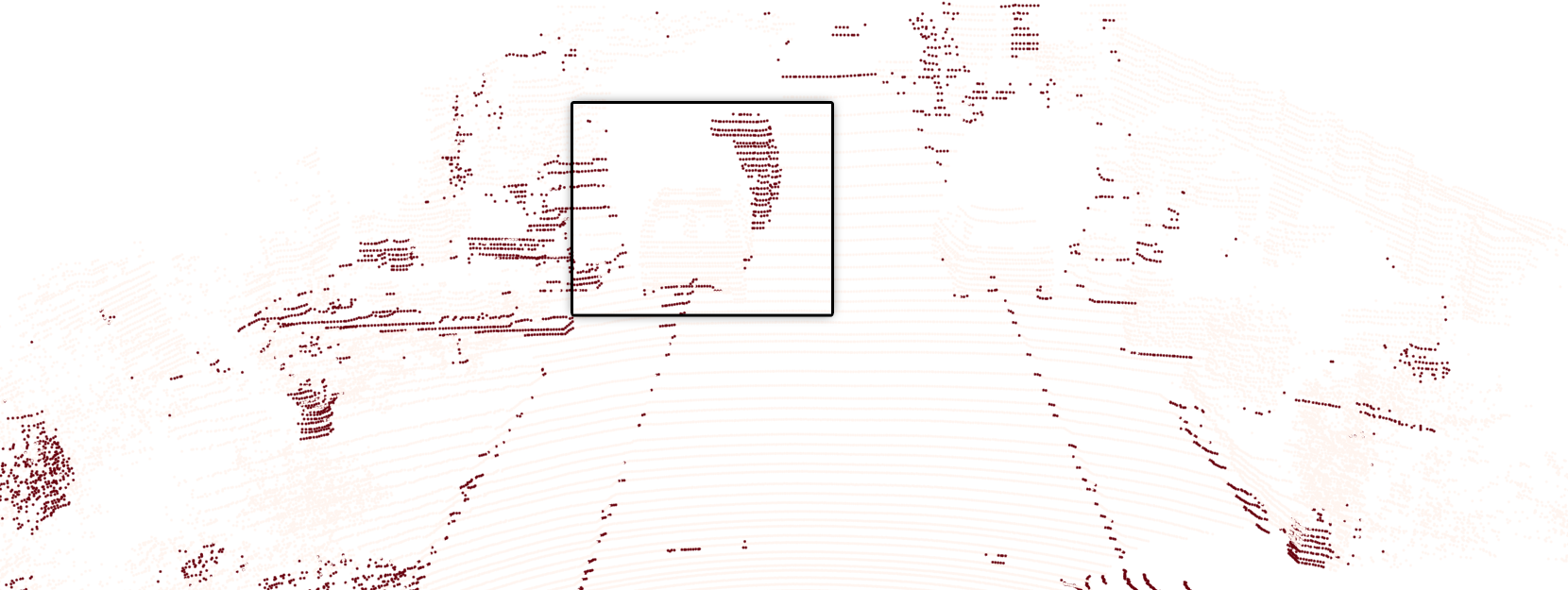}} \\
\\
\rotatebox[origin=c]{90}{(b)} & {\includegraphics[width=\linewidth, frame]{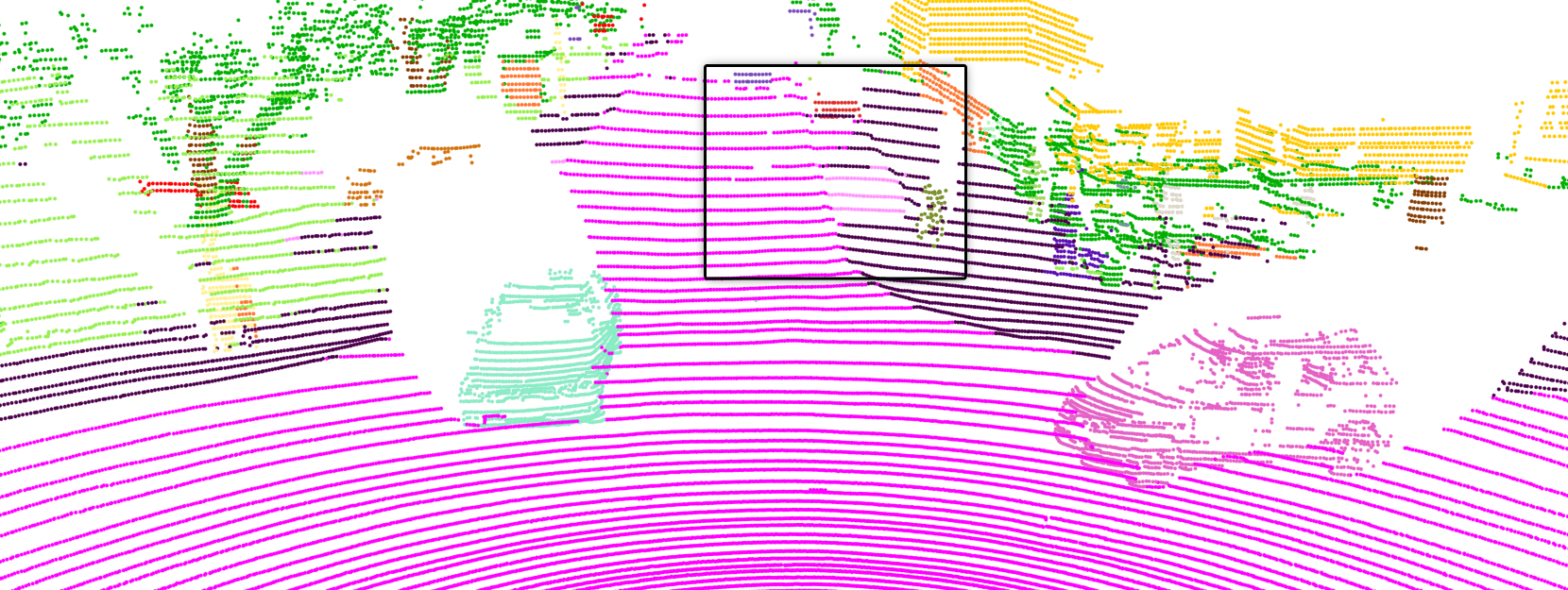}} & {\includegraphics[width=\linewidth, frame]{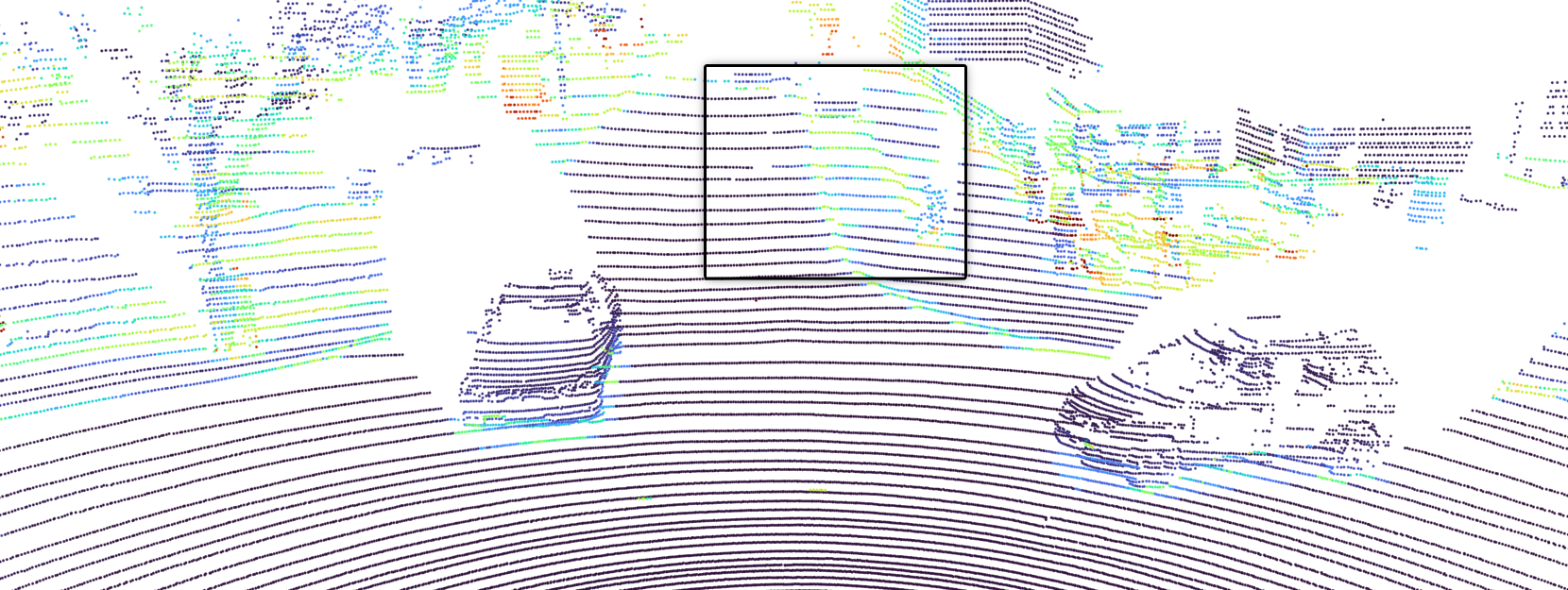}} & {\includegraphics[width=\linewidth, frame]{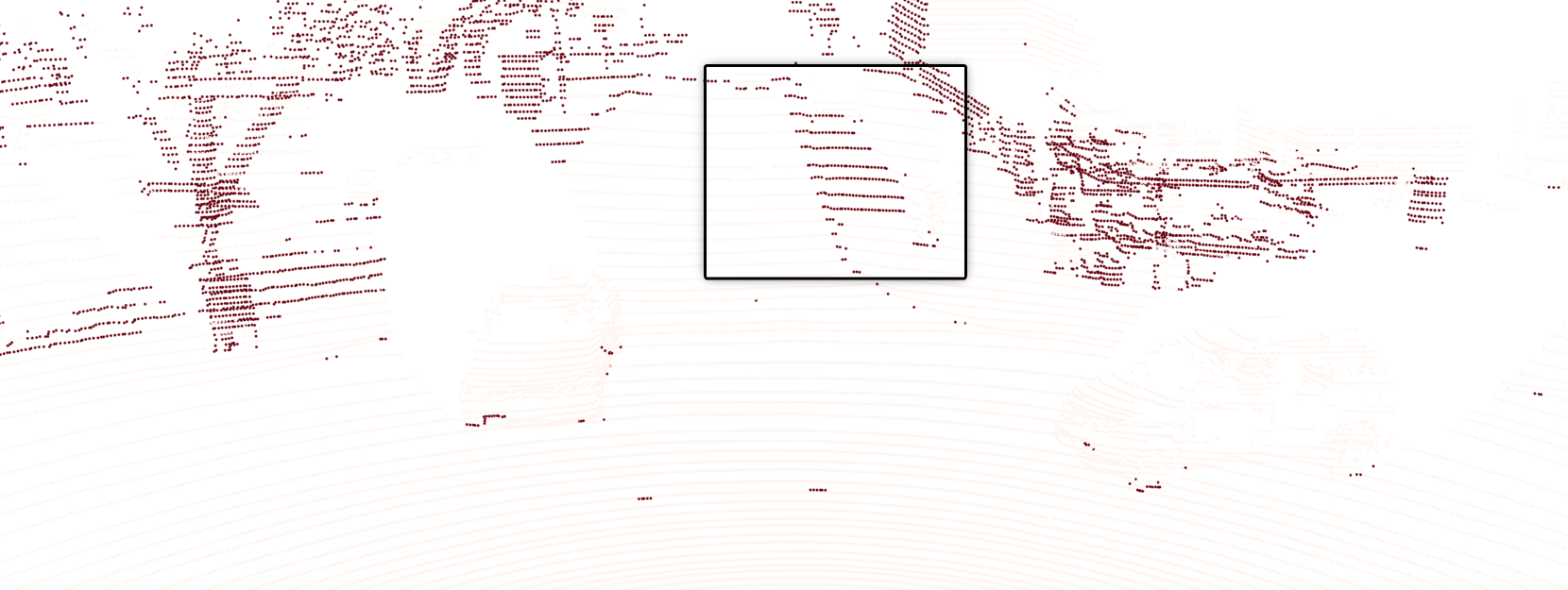}} \\
\\
\end{tabular}
}
\caption{Qualitative results for the uncertainty-aware panoptic segmentation by our EvLPSNet for two scans of the SemanticKITTI validation set.
Red regions in the uncertainty map depict high predicted uncertainty, and dark regions in the error map depict misclassified points.
}
\label{fig:qualitative results}
\vspace{-0.5ex}
\end{figure*}

\subsection{Quantitative Results}

The experimental results on the SemanticKITTI validation dataset are presented in \tabref{tab:quant_val}.
Our proposed EvLPSNet achieves the best score for the overall uPQ, signifying its superior performance for the uncertainty-aware panoptic segmentation task,
as well for the pECE and uECE, signifying the most accurate uncertainty calibration of all methods.
Interestingly, our network is also superior to the Panoptic-PolarNet in PQ$^\text{st}$ and mIoU,
which we attribute to our proposed uQR and pKNN modules for refining segmentation results utilizing the uncertainties and probabilities, see \secref{sec:ablation}.
Comparing the overall PQ, our approach is slightly worse than Panoptic-PolarNet.
However, given its well calibrated uncertainty estimation, our network recovers the performance on uPQ.
We observe mixed results for temperature scaling (TS). It generally improves uECE, but pECE is only improved for EfficientLPS, while it is worsened for Panoptic-PolarNet.
This is similar to \cite{sirohi2022uncertainty}, where TS was able to improve uECE, but the advantage almost vanishes for the panoptic pECE.

\subsection{Qualitative Results}

We present qualitative results including the predicted panoptic segmentation and uncertainties together with the error maps in \figref{fig:qualitative results}.
For example, in \figref{fig:qualitative results}a, we depict a misclassified vehicle (marked by the box).
We observe that the wrongly classified points (see error map) are well represented with high predicted uncertainties in the network output.
Similarly, in \figref{fig:qualitative results}b, a sidewalk (dark magenta) is wrongly predicted as drivable road (light pink), but high uncertainties are predicted here as well.
Generally, the uncertainty prediction is strongly correlated with the error map, validating the prediction quality.

\begin{table*}
\begin{center}
\rule{0pt}{0.5ex}    
\caption{Class-wise PQ values in \% on the SemanticKITTI validation set. Lower values are better for $\downarrow$, and larger values otherwise.}
\label{tab:ablations}
\footnotesize
\begin{tabular}
{l|cc|ccccccccc|cc|cc|c}
\toprule
Model & pKNN & uQR & \begin{sideways}car\end{sideways} & \begin{sideways}bicycle\end{sideways}  &
\begin{sideways}m.cycle\end{sideways} & \begin{sideways}person\end{sideways} & \begin{sideways}bicyclist\end{sideways} & \begin{sideways}m.cyclist\end{sideways}& \begin{sideways}trunk\end{sideways} & 
\begin{sideways}pole\end{sideways} &
\begin{sideways}tr. sign\end{sideways} & uPQ & PQ & pECE $\downarrow$ & uECE $\downarrow$ & mIoU\\
\midrule
M1 & \xmark  & \xmark & $87.9$ & $51.8$ & $59.0$ & $55.0$ & $83.5$ & $44.1$ & $45.5$ & $55.9$ & $48.6$ & $51.4$ & $57.3$ & $\textbf{10.4}$ & $\textbf{2.7}$ &$63.5$   \\
M2 & \cmark & \xmark &$88.0$ & $52.1$ & $60.0$ & $57.3$ & $84.1$ & $\textbf{44.7}$ & $45.9$ & $56.0$ & $49.0$ & $\textbf{51.5}$ & $57.7$ & $10.6$ & $2.8$& $63.7$   \\
M3 & \xmark & \cmark &$\textbf{88.1}$ & $53.3$ & $59.6$ & $55.3$ & $84.1$ & $44.2$ & $47.8$ & $\textbf{56.3}$ & $49.9$ & $51.2$ & $57.8$ & $11.3$ & $3.9$ & $63.9$   \\
M4 & \cmark & \cmark &$\textbf{88.1}$ & $\textbf{53.4}$ & $\textbf{60.5}$ & $\textbf{57.5}$ & $\textbf{84.6}$ & $\textbf{44.7}$ & $\textbf{47.9}$ & $56.2$ & $\textbf{49.9}$ & $51.4$ & $\textbf{58.0}$ & $11.5$ & $7.1$& $\textbf{64.0}$\\
\bottomrule
\end{tabular}
\end{center}
\end{table*}

\begin{figure}
\vspace{-0.5ex}
\centering
\includegraphics[width=0.9\linewidth]{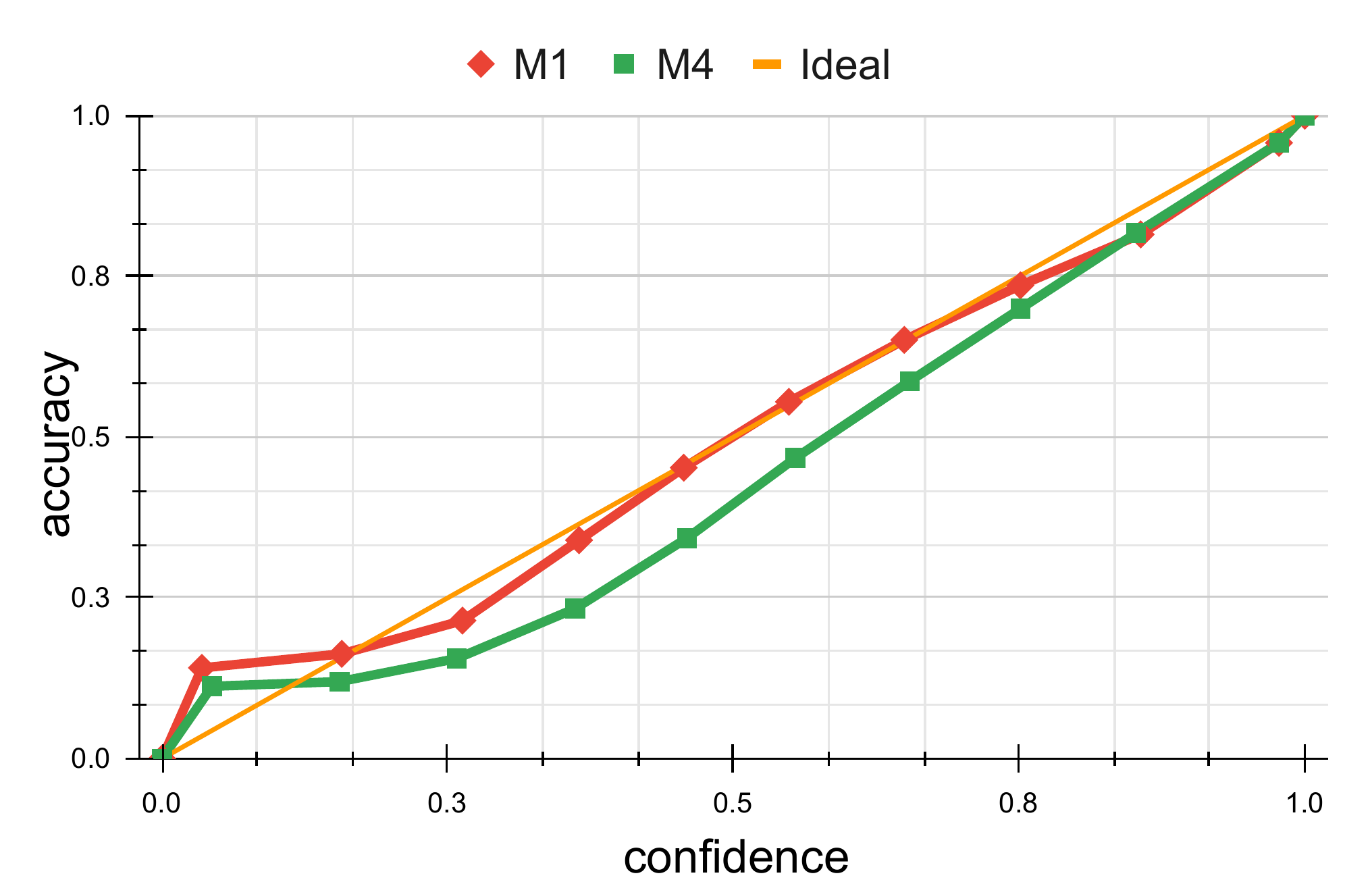}
\caption{Calibration curves for our M1 and M4 models.} 
\label{fig:calibrationcurve}
\vspace{-0.5ex}
\end{figure}

\subsection{Ablation Studies}\label{sec:ablation}

\subsubsection{Refinement Modules}

We present the quantitative analysis of our proposed uQR and pKNN modules in \tabref{tab:ablations} for the overall performance and for selected classes.
The motivation behind both modules is to utilize uncertainties to improve the segmentation performance, specifically of points suffering from discretization errors.
This affects in particular smaller objects that consist of fewer points, such as persons, poles, and traffic signs.

Model M1 leaves out the pKNN and uQR modules compared to our final approach.
In model M2, we incorporate the pKNN module, which leads to a gain in PQ for all mentioned classes, with the most gain for the person class.
The overall uPQ, PQ and mIoU are slightly improved as well, however, pECE and uECE are slightly worsened.
Similarly, in model M3 we incorporate the uQR module, improving PQ and mIoU, but slightly worsening uPQ, pECE and uECE.
Both results, for M2 and M3, signify a loss in calibration quality, but a gain in segmentation performance.
Model M4 is our final model, incorporating both the pKNN and uQR modules.
We observe that the contributions from both modules roughly add up, in particular a significant gain in PQ for most of the small classes can be seen.

We further present calibration curves for the models M1 and M4 in \figref{fig:calibrationcurve}. 
It can be observed that the application of our refinement modules renders the network slightly overconfident.
In conclusion, our final M4 model shows the best segmentation performance, however,
our M1 model shows the most accurate uncertainty calibration. 

\begin{figure}
\vspace{-0.5ex}
\centering
\includegraphics[width=0.9\linewidth]{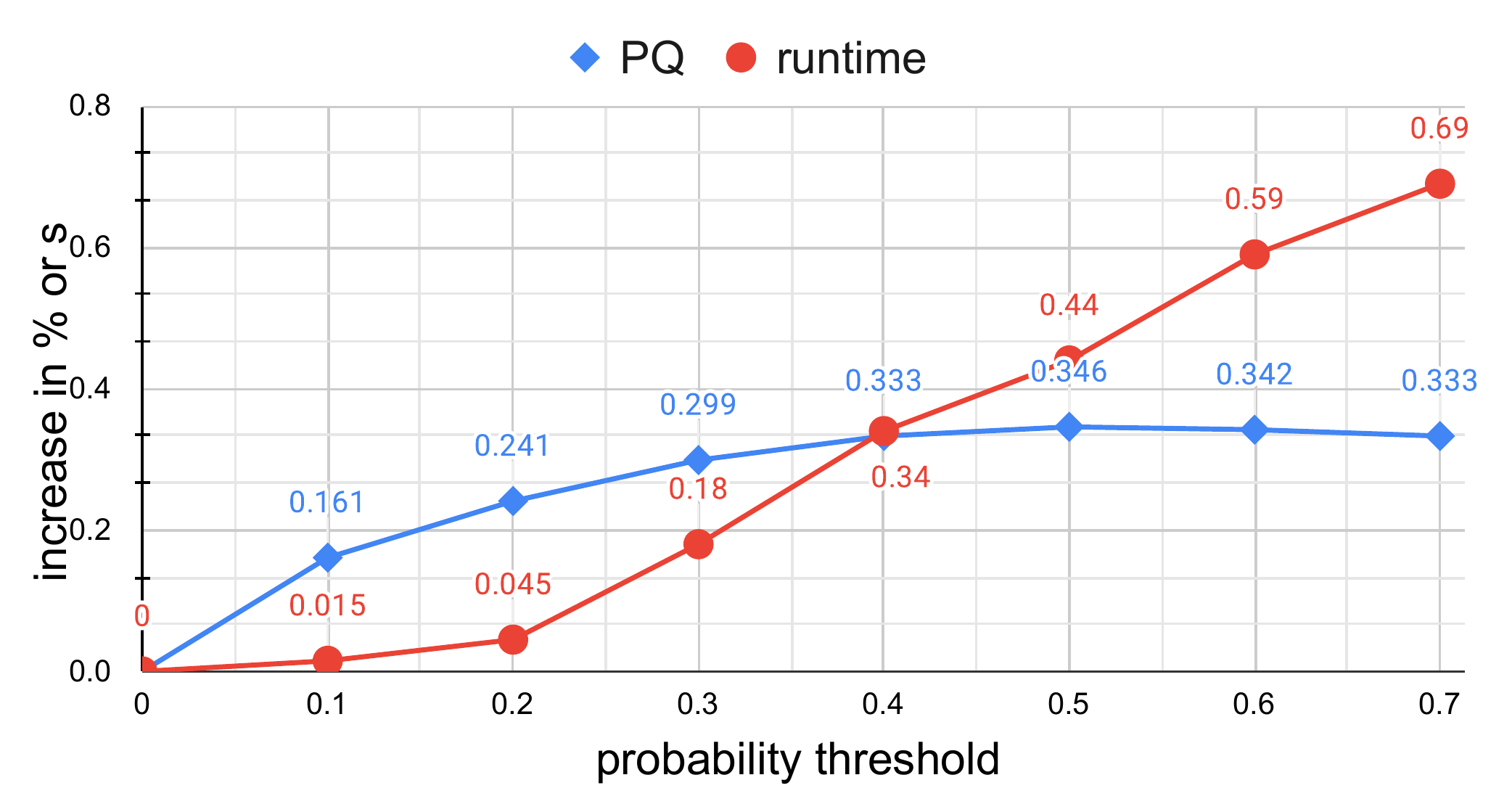}
\caption{PQ gain and runtime versus pKNN probability threshold.} 
\label{fig:pknncurve}
\vspace{-0.5ex}
\end{figure}

\subsubsection{Efficiency of pKNN}

The efficiency of our pKNN module is showcased in \figref{fig:pknncurve}.
In our experiment, we increment the probability threshold that is used to select the points for applying the KNN post-processing. We report the increase in PQ and in runtime of the network.
Here, we used a subset of the validation set.
The results compare to a base runtime of about 0.09 seconds per frame without pKNN.
We observe that the PQ and runtime increase with the threshold, which is expected due to an increase in the number of selected points.
However, the PQ saturates for a threshold of about 0.4, our final value.
For a faster evaluation one can lower the threshold to e.g. 0.2 for a modest performance gain with low increase in runtime.

\section{Conclusions}

In this work we proposed EvLPSNet, a novel proposal-free approach for solving the task of uncertainty-aware LiDAR panoptic segmentation.
It is the first network to simultaneously predict panoptic segmentation and uncertainties of LiDAR point clouds in a single forward pass.
To this end, we demonstrated an effective way of utilizing evidential deep learning for our uncertainty-aware semantic segmentation head.
We further proposed an uncertainty-based query and refining (uQR) module to leverage and improve the segmentation of points that suffer from discretization errors.
Moreover, our pKNN module showcased how probabilities can be beneficial to reduce the runtime of KNN clustering methods, while maintaining the gain in performance.
Our network achieves the best performance on the uncertainty-aware panoptic segmentation performance and calibration metrics, uPQ and pECE, respectively.
We hope our work will motivate future works in holistic and reliable 3D scene understanding using LiDAR point clouds.

\bibliographystyle{IEEEtran}
\bibliography{main.bib}

\end{document}